%% file: main.tex
\pdfoutput=1
\documentclass{article} 
\usepackage{iclr2026_conference,times}

\input{math_commands.tex}

\usepackage[utf8]{inputenc}
\usepackage[T1]{fontenc}
\usepackage{hyperref}
\usepackage{url}
\usepackage{amsmath,amssymb,amsthm,mathtools}
\usepackage{booktabs}
\usepackage{graphicx}
\usepackage{microtype}
\usepackage{xcolor}
\usepackage[capitalise,noabbrev]{cleveref}

\hypersetup{colorlinks=true,linkcolor={blue!60!black},citecolor={blue!60!black},urlcolor={blue!60!black}}

\newtheorem{theorem}{Theorem}
\newtheorem{proposition}{Proposition}
\newtheorem{corollary}{Corollary}
\newtheorem{lemma}{Lemma}
\theoremstyle{definition}
\newtheorem{definition}{Definition}
\newtheorem{remark}{Remark}
\crefname{lemma}{Lemma}{Lemmas}

\newcommand{\sig}{\mathsf{s}}               

\newcommand{\dd}{\mathrm{d}}

\renewcommand{\vf}{v}                           
\newcommand{\bfn}{b}                          
\newcommand{\Rde}{\mathcal{R}}                
\newcommand{\Wone}{\overline{W}_{1}}

\newcommand{\Ep}{\mathbb{E}_{\pi}}
\newcommand{\op}{\mathrm{op}}

\title{A Transport Geometry of Benign Overfitting}

\author{Olaf Yunus Laitinen Imanov\textsuperscript{$\ast\dagger$} \\
Department of Mathematics and Statistics\\
University of Helsinki, Finland\\
\texttt{yunus.z.imanov@helsinki.fi}
\And
Rana Irem Turhan\textsuperscript{$\ast$} \\
School of Computer Science\\
University of Sheffield, United Kingdom\\
\texttt{riturhan1@sheffield.ac.uk}
\AND
Isgandar Jafarli\textsuperscript{$\ast$} \\
School of IT and Engineering\\
ADA University, Azerbaijan\\
\texttt{ijafarli14955@ada.edu.az}
}

\iclrfinalcopy
\begin{document}

\maketitle
\lhead{Preprint}
{\renewcommand{\thefootnote}{}\footnotetext{\textsuperscript{$\ast$}Equal contribution. \textsuperscript{$\dagger$}Corresponding author.}}

\begin{abstract}
Benign overfitting, the phenomenon that interpolating noisy data can be statistically harmless, is usually characterized through effective-rank conditions on the covariance spectrum. These conditions are fragile. We show that a perturbation of vanishing relative operator and Frobenius norm can turn a benign covariance into one whose minimum-norm interpolator has non-vanishing excess risk. We argue that the natural geometry of benign overfitting is not that of the covariance matrix but that of \emph{optimal transport of its log-eigenvalues, measured in units of the sample size}. In log-spectral coordinates, the deterministic equivalent of the risk of minimum-norm interpolation becomes a functional of two measures smoothed by a logistic kernel centred at the interpolation threshold $\log\kappa$. We compute the exact first variation of this functional. The threshold moves by a sensitivity-weighted average of the log-eigenvalue displacements, and the influence of each eigenvalue decays exponentially with its log-distance to the threshold. We then derive global, dimension-free stability bounds. The variance functional is $\frac{8}{27n}$-Lipschitz with respect to the $1$-Wasserstein distance between log-spectra, with a sharp constant, and is controlled more generally by a threshold-localized transport discrepancy. An analogous localized bound controls the bias through the cumulative signal measure. As consequences, any rank-$k$ update of the covariance, of arbitrary norm, changes the variance functional by at most $3k/n$, and benign overfitting survives every perturbation whose localized transport cost is $o(n)$. The results extend verbatim to ridge regression and to infinite-dimensional trace-class covariances. Matching lower bounds show that the $n$-scaling is necessary and cannot be replaced by operator norms, Frobenius norms, or dimension-normalized spectral distances. Simulations, one-step feature learning in two-layer networks, and random-feature representations of real images support the theory and delineate its limits.
\end{abstract}

\section{Introduction}
\label{sec:intro}

Modern predictors routinely interpolate their training data, noise included, and still generalize \citep{zhang2017understanding,belkin2019reconciling,nakkiran2020deep}. The cleanest setting in which this \emph{benign overfitting} is understood is linear regression with the minimum-$\ell_2$-norm interpolator. There, \citet{bartlett2020benign} showed that the excess risk vanishes if and only if the covariance spectrum satisfies two effective-rank conditions, and a large body of work has refined and extended this picture \citep{muthukumar2020harmless,belkin2020two,hastie2022surprises,koehler2021uniform,tsigler2023benign,mallinar2022benign}.

\begin{figure}[t]
\centering
\includegraphics[width=\linewidth]{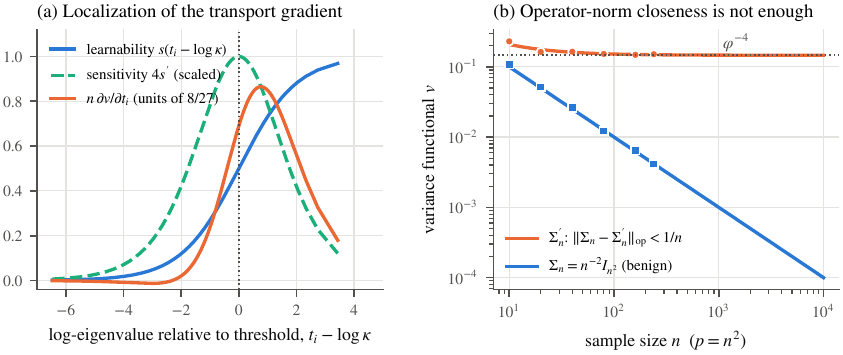}
\vspace{-1.5em}
\caption{\textbf{(a)} For $\lambda_i=1/i$ ($p=2\cdot 10^4$, $n=200$), the exact gradient of the variance functional with respect to the log-eigenvalues (\cref{thm:firstvar}) is concentrated in an $O(1)$ window around the interpolation threshold $\log\kappa$; directions far above or far below the threshold are irrelevant. \textbf{(b)} A perturbation with $\|\Sigma_n-\Sigma_n'\|_{\op}<1/n$ (here shown without the unit spike of \cref{thm:lower}, which changes $v$ by at most $3/n$) turns benign overfitting ($v=1/n$) into tempered overfitting ($v\to\varphi^{-4}$). Lines: deterministic equivalent; markers: simulation of the minimum-norm interpolator ($30$ draws per point; markers show $\bar V/(\sigma^2+\bar V)$ for the simulated variance $\bar V$).}
\label{fig:intro}
\vspace{-0.8em}
\end{figure}

Effective-rank conditions are statements about one covariance matrix. In practice the relevant covariance is never known exactly. It is the covariance of a learned representation that changes during training, of features computed on shifted or augmented data, or of an estimate from finite samples. This raises a basic question that, to our knowledge, has not been answered quantitatively:
\begin{center}
\emph{Which perturbations of the covariance preserve benign overfitting, and in what geometry is it continuous?}
\end{center}
The obvious candidates fail. \Cref{thm:lower}(a) exhibits covariances $\Sigma_n,\Sigma_n'$ with $\|\Sigma_n-\Sigma_n'\|_{\op}/\|\Sigma_n\|_{\op}\to 0$ and $\|\Sigma_n-\Sigma_n'\|_{F}/\|\Sigma_n\|_{F}\to0$, whose empirical spectral distributions also converge to each other in Kolmogorov and Wasserstein distance. Yet the variance of the minimum-norm interpolator vanishes under $\Sigma_n$ and converges to $\sigma^2/(\varphi^4-1)\approx 0.17\,\sigma^2$ under $\Sigma_n'$, where $\varphi$ is the golden ratio (\cref{fig:intro}b). Conversely, adding spikes of operator norm $10^6$ to a benign covariance changes the variance only by an amount proportional to the number of spikes (\cref{fig:main}a). Operator-norm closeness is thus neither sufficient nor necessary.

\paragraph{Contributions.} We show that the right geometry is optimal transport of \emph{log}-eigenvalues, with costs counted in units of the \emph{sample size} $n$ rather than the dimension.
\begin{enumerate}
\item \textbf{Log-spectral representation (\cref{sec:logspec}).} In the coordinates $t_i=\log\lambda_i$, the standard deterministic equivalent of the risk of minimum-norm interpolation \citep{hastie2022surprises,bach2024high,cheng2024dimension,misiakiewicz2024nonasymptotic} is a functional of the spectral counting measure and a signal measure, both convolved with a logistic kernel centred at the interpolation threshold $u=\log\kappa$. The threshold is the point at which the smoothed counting measure has mass exactly $n$, which is the conservation law of \citet{simon2023eigenlearning} in transport language.
\item \textbf{Exact first variation (\cref{thm:firstvar}).} Under an infinitesimal transport of log-eigenvalues, the threshold moves by the average displacement under a \emph{sensitivity distribution} $\pi_i\propto \lambda_i\kappa/(\lambda_i+\kappa)^2$. Every component of the gradient of the risk is proportional to $\pi_i$, so the influence of an eigenvalue decays like $e^{-|\log(\lambda_i/\kappa)|}$.
\item \textbf{Global stability (\cref{cor:w1,thm:local,prop:winf}).} The variance functional $v$, which decides benign versus tempered versus catastrophic overfitting of noise, satisfies $|v(\Sigma)-v(\Sigma')|\le \frac{8}{27}\,\Wone/n$ with a sharp constant, where $\Wone$ is the translation-reduced $1$-Wasserstein distance between log-spectra. A localized refinement weighs counting-function discrepancies by a kernel of total mass $3$ centred at $\log\kappa$. A companion bound controls the bias through the cumulative signal measure (\cref{thm:bias}). Corollaries: every rank-$k$ update of arbitrary norm moves $v$ by at most $3k/n$, and benign overfitting (variance \emph{and} bias) is preserved under any perturbation whose localized transport cost is $o(n)$. All of this holds for ridge regression and for infinite-dimensional trace-class covariances (\cref{prop:ext}).
\item \textbf{Necessity (\cref{thm:lower}).} The $n$-scaling cannot be removed: operator and Frobenius norms (absolute or relative) and dimension-normalized spectral distances do not control $v$, and the bound $3k/n$ is attained up to a factor $3$.
\item \textbf{Experiments (\cref{sec:exp}).} Simulations confirm all predictions. In one-step feature learning \citep{ba2022high}, the representation's covariance changes by more than $100\times$ in operator norm, while $v$ moves by less than $0.04$, as the transport certificate predicts. On random features of real images, the theory is accurate wherever the deterministic equivalent is, and we document one structured perturbation where any spectrum-only theory must fail.
\end{enumerate}

\section{Setting and background}
\label{sec:setting}

We observe $n$ i.i.d.\ pairs $(x_j,y_j)$ with $x_j\in\R^p$, $\E x_j=0$, $\E x_jx_j^\top=\Sigma$, and $y_j=\langle x_j,\theta^\star\rangle+\xi_j$, where the noise $\xi_j$ is independent with mean zero and variance $\sigma^2$. Let $X\in\R^{n\times p}$ be the design matrix with $\mathrm{rank}(\Sigma)>n$. The minimum-norm interpolator is $\hat\theta=X^{+}y$, and its excess risk is $\|\hat\theta-\theta^\star\|_\Sigma^2=\E_x\langle x,\hat\theta-\theta^\star\rangle^2$. Write $\Sigma=\sum_i\lambda_ie_ie_i^\top$ with $\lambda_1\ge\lambda_2\ge\dots$, and let $\theta_i=\langle\theta^\star,e_i\rangle$. Only the positive eigenvalues matter below, so we index $i=1,\dots,p$ over them.

\paragraph{The deterministic equivalent.} Let $\kappa>0$ be the unique solution of
\begin{equation}\label{eq:kappa}
\textstyle\sum_{i}\frac{\lambda_i}{\lambda_i+\kappa}=n ,
\end{equation}
which exists because the left-hand side decreases continuously from $p>n$ to $0$. Let $\mathrm{df}_2=\sum_i\lambda_i^2/(\lambda_i+\kappa)^2$. A long line of work \citep{dobriban2018high,hastie2022surprises,bach2024high,cheng2024dimension,misiakiewicz2024nonasymptotic,atanasov2024scaling} shows that, under suitable conditions on the design (e.g.\ Gaussian or with concentration properties), the excess risk of $\hat\theta$ is approximated, with explicit non-asymptotic error bounds in the cited works, by
\begin{equation}\label{eq:de}
\Rde_n(\Sigma,\theta^\star,\sigma^2)\;=\;\underbrace{\frac{\kappa^2\,\langle\theta^\star,\Sigma(\Sigma+\kappa)^{-2}\theta^\star\rangle}{1-\mathrm{df}_2/n}}_{\text{bias } B}\;+\;\underbrace{\frac{\sigma^2\,\mathrm{df}_2}{n-\mathrm{df}_2}}_{\text{variance } V}.
\end{equation}
The same object governs kernel ridge(less) regression and random-feature models through their spectra \citep{bordelon2020spectrum,canatar2021spectral,jacot2020kernel,simon2023eigenlearning,defilippis2024dimension}. \textbf{Throughout, our theorems are exact statements about the functional $\Rde_n$.} They transfer to the actual risk whenever an approximation theorem of the above type applies, at the cost of that theorem's additive error. \Cref{rem:transfer} makes this explicit. We do not re-derive the approximation theorems themselves. Instead, we verify numerically that the approximation holds in our experiments, for Gaussian and Rademacher designs and with and without ridge penalty (\cref{tab:val,tab:val2} in \cref{app:exp}).

\paragraph{The variance functional and the taxonomy of overfitting.} Define $v=\mathrm{df}_2/n\in(0,1)$. Then $V=\sigma^2 v/(1-v)$. In particular, the noise-fitting part of the risk is a monotone function of $v$ alone.
\begin{definition}\label{def:regimes}
A sequence of problems indexed by $n$ is \emph{variance-benign} if $v_n\to0$, \emph{tempered} if $v_n\to v_\infty\in(0,1)$, and \emph{catastrophic} if $v_n\to1$.
\end{definition}
This mirrors the taxonomy of \citet{mallinar2022benign}. \Cref{lem:effrank} in \cref{app:proofs} shows that $v$ is comparable, up to absolute constants, to $k/n+n\sum_{i>k}\lambda_i^2/(\sum_{i>k}\lambda_i)^2$ with $k=\#\{i:\lambda_i\ge\kappa\}$ whenever $k\le n/2$. This is the effective-rank expression of \citet{bartlett2020benign}. Benign overfitting additionally requires the bias to vanish. We treat the bias in \cref{thm:bias,prop:winf}.

\section{The log-spectral picture}
\label{sec:logspec}

Let $\sig(x)=1/(1+e^{-x})$ be the logistic function, and let $t_i=\log\lambda_i$ and $u=\log\kappa$. Since $\lambda/(\lambda+\kappa)=\sig(\log\lambda-\log\kappa)$, everything in \eqref{eq:de} is expressed through
\[
\sig_i:=\sig(t_i-u)\in(0,1),\qquad \sig_i':=\sig_i(1-\sig_i),\qquad m_i:=\lambda_i\theta_i^2\ \ (\text{signal energy in direction } i).
\]
We encode the problem by two measures on $\R$: the \emph{spectral counting measure} $\mu=\sum_i\delta_{t_i}$ and the \emph{signal measure} $\nu=\sum_i m_i\delta_{t_i}$, with $\nu(\R)=\|\theta^\star\|_\Sigma^2$. We write $N_\mu(t)=\mu((t,\infty))$ for the counting function, so that $N_\mu(t)$ is the number of eigenvalues larger than $e^{t}$.

\begin{proposition}[Log-spectral representation]\label{prop:logspec}
Let $\mu$ have $p>n$ atoms. Then:
\begin{enumerate}
\item[(i)] $u=\log\kappa$ is the unique solution of $\int\sig(t-u)\,\dd\mu(t)=n$. Equivalently, $P_\mu:=\frac1n\sum_i\sig_i\delta_{t_i}$ is a probability measure.
\item[(ii)] $\vf=\frac1n\sum_i\sig_i^2$, $\ \bfn:=\kappa^2\langle\theta^\star,\Sigma(\Sigma+\kappa)^{-2}\theta^\star\rangle=\int(1-\sig(t-u))^2\,\dd\nu(t)$, and $\Rde_n=(\sigma^2\vf+\bfn)/(1-\vf)$.
\item[(iii)] (Translation invariance) Shifting $\mu$ and $\nu$ by $c\in\R$, i.e.\ $\Sigma\mapsto e^{c}\Sigma$ with $\theta^\star\mapsto e^{-c/2}\theta^\star$, shifts $u$ by $c$ and leaves $\vf,\bfn,\Rde_n$ unchanged.
\item[(iv)] (Counting-function form) For every $w\in\R$, $\int\sig(t-w)\,\dd\mu=\int N_\mu(t)\,\sig'(t-w)\,\dd t$ and $\int\sig(t-w)^2\,\dd\mu=\int N_\mu(t)\,(\sig^2)'(t-w)\,\dd t$.
\end{enumerate}
\end{proposition}
Part (i) is the conservation law $\sum_i\mathcal L_i=n$ for the learnabilities $\mathcal L_i=\sig_i$ of \citet{simon2023eigenlearning}. The novelty lies in reading it geometrically. The threshold $u$ is a \emph{location parameter} of the log-spectrum, and every functional in \eqref{eq:de} is a logistic smoothing of $\mu$ or $\nu$ at unit scale around it. Three probability distributions on directions will appear repeatedly:
\begin{equation}\label{eq:dists}
\underbrace{\pi_i=\frac{\sig_i'}{\sum_j\sig_j'}}_{\text{sensitivity}},\qquad
\underbrace{w_i=\frac{m_i(1-\sig_i)^2}{\bfn}}_{\text{bias share}},\qquad
\underbrace{P_\mu(\{t_i\})=\frac{\sig_i}{n}}_{\text{interpolation measure}}.
\end{equation}
Note that $\sum_j\sig_j'=n-\sum_j\sig_j^2=n(1-\vf)$, and that $\sig'(x)=\frac{1}{4\cosh^2(x/2)}\le e^{-|x|}$. Hence $\pi$ is concentrated on the directions within $O(1)$ of the threshold on the log scale.

\section{First variation: the transport gradient}
\label{sec:firstvar}

We first ask how the risk responds when the log-eigenvalues are transported infinitesimally.

\begin{theorem}[Transport gradient]\label{thm:firstvar}
Let $p>n$ be finite. The maps $(t,m)\mapsto u,\vf,\bfn,\Rde_n$ are real-analytic on $\R^p\times(0,\infty)^p$, and for every $j$:
\begin{align*}
\text{(a)}\quad &\frac{\partial u}{\partial t_j}=\pi_j,
&\text{(b)}\quad &\frac{\partial \vf}{\partial t_j}=\frac{2}{n}\,\sig_j'\big(\sig_j-\Ep[\sig]\big),\\
\text{(c)}\quad &\frac{\partial \bfn}{\partial t_j}=2\,\sig_j'\big(\Ep[m(1-\sig)]-m_j(1-\sig_j)\big),
&\text{(d)}\quad &\frac{\partial \bfn}{\partial \log m_j}=m_j(1-\sig_j)^2,
\end{align*}
where $\Ep[f]=\sum_i\pi_if_i$. Moreover, $\partial\Rde_n/\partial t_j=\big[(\sigma^2+\Rde_n)\,\partial_{t_j}\vf+\partial_{t_j}\bfn\big]/(1-\vf)$, $\sum_j\partial_{t_j}u=1$, $\sum_j\partial_{t_j}\vf=\sum_j\partial_{t_j}\bfn=0$, and
\begin{equation}\label{eq:gradbound}
\Big|\frac{\partial \vf}{\partial t_j}\Big|\le\frac{2}{n}\,\sig_j'\le\frac{2}{n}\,e^{-|t_j-u|},\qquad \sup_{j}\Big|\frac{\partial \vf}{\partial t_j}\Big|\le\frac{8}{27\,n}.
\end{equation}
\end{theorem}

\paragraph{Reading the theorem.} Part (a) says that $\dd\log\kappa=\Ep[\dd\log\lambda]$: the threshold moves by the sensitivity-weighted mean displacement. Part (b) says that pushing eigenvalue $j$ up increases the variance exactly when its learnability exceeds the sensitivity-weighted mean learnability $\Ep[\sig]$. Every displacement gradient carries the factor $\sig_j'$, so eigenvalues far above the threshold (already fitted, $\sig_j\approx1$) and far below it (ignored, $\sig_j\approx0$) have exponentially small influence (\cref{fig:intro}a). The identities $\sum_j\partial_{t_j}\vf=0$ express scale invariance: only the \emph{shape} of the log-spectrum matters. This is why all our distances below are reduced modulo translations.

Integrating \eqref{eq:gradbound} along straight-line transport paths yields a global bound with a sharp constant. For $\mu,\mu'$ with the same number $p$ of atoms, let $t_{(1)}\le\dots\le t_{(p)}$ and $t'_{(1)}\le\dots\le t'_{(p)}$, and define the translation-reduced (unnormalized) $1$-Wasserstein distance
\[
\Wone(\mu,\mu'):=\inf_{c\in\R}\sum_{i=1}^{p}\big|t'_{(i)}-c-t_{(i)}\big|=\inf_{c\in\R}\int_\R\big|N_\mu(t)-N_{\mu'}(t+c)\big|\,\dd t .
\]
The two expressions agree because monotone matching is optimal in one dimension \citep{villani2009optimal,santambrogio2015optimal}.

\begin{corollary}[Sharp Wasserstein--Lipschitz bound]\label{cor:w1}
For all $n$ and all $\mu,\mu'$ with the same number $p>n$ of atoms, $|\vf(\mu)-\vf(\mu')|\le\frac{8}{27}\,\Wone(\mu,\mu')/n$. The constant is sharp: $\sup n\,|\vf(\mu)-\vf(\mu')|/\Wone(\mu,\mu')=8/27$, where the supremum ranges over $n$ and all such pairs.
\end{corollary}

The normalization by $n$, not by $p$, is the first structural message. Transporting $o(n)$ units of log-eigenvalue mass cannot change the variance functional by more than $o(1)$, no matter how large the dimension is. The bound is still blind to localization: moving $p\gg n$ tail eigenvalues by a small amount costs $\Theta(p)$, while \cref{thm:firstvar} shows their influence is negligible. The next result repairs this.

\section{Global stability in a threshold-localized transport geometry}
\label{sec:global}

For two counting measures $\mu,\mu'$, possibly with different numbers of atoms $p,p'>n$, a shift $c\in\R$, and $u=\log\kappa(\mu)$, define the \emph{localized discrepancies}
\begin{equation}\label{eq:localdisc}
\Lambda_c:=\int_\R\big|N_\mu(t)-N_{\mu'}(t+c)\big|\,\sig'(t-u)\,\dd t,\qquad
\Psi_c:=\int_\R\big|N_\mu(t)-N_{\mu'}(t+c)\big|\,\psi(t-u)\,\dd t,
\end{equation}
with $\psi=2\sig'(1+\sig)$. The kernels have total mass $\int\sig'=1$ and $\int\psi=3$, peak values $\frac14$ and $\frac{4}{3\sqrt3}$, and both decay like $e^{-|t-u|}$. The quantity $\Psi_c$ is a transport cost between log-spectra in which mass is charged only to the extent that it crosses levels near the interpolation threshold.

\begin{theorem}[Localized transport stability]\label{thm:local}
Let $\vf=\vf(\mu)$ and $\vf'=\vf(\mu')$. For every $c\in\R$:
\begin{enumerate}
\item[(i)] $|\vf'-\vf|\le\Psi_c/n$.
\item[(ii)] If $\Lambda_c\le n(1-\vf)/6$, then $\big|\log\kappa(\mu')-c-\log\kappa(\mu)\big|\le 6\Lambda_c/\big(n(1-\vf)\big)$.
\item[(iii)] $\Psi_c\le3\,\sup_t|N_\mu(t)-N_{\mu'}(t+c)|$ and $\Lambda_c\le\sup_t|N_\mu(t)-N_{\mu'}(t+c)|$. If $p=p'$, then also $\Psi_c\le\frac{4}{3\sqrt3}\int|N_\mu(t)-N_{\mu'}(t+c)|\,\dd t$ and $\Lambda_c\le\frac14\int|N_\mu(t)-N_{\mu'}(t+c)|\,\dd t$.
\end{enumerate}
\end{theorem}

The proof (\cref{app:proofs}) avoids paths altogether. By \cref{prop:logspec}(iv), both $\sum_i\sig_i$ and $\sum_i\sig_i^2$ are integrals of the counting function against fixed kernels. The change in $\vf$ splits into a change at a fixed threshold, controlled by $\int|N-N'|(\sig^2)'$, and a change caused by moving the threshold. The latter is controlled by the change of $\sum_i\sig_i$, since $|\frac{\dd}{\dd w}\sum\sig^2|\le2|\frac{\dd}{\dd w}\sum\sig|$. The statement holds for different dimensions $p\ne p'$ and is dimension-free.

\begin{corollary}[Benign overfitting is continuous in the transport geometry]\label{cor:benign}
Let $\Sigma_n,\Sigma_n'$ be sequences of covariances with log-spectral counting measures $\mu_n,\mu_n'$. If $\vf(\Sigma_n)\to0$ and $\inf_c\Psi_c(\mu_n,\mu_n')=o(n)$, then $\vf(\Sigma_n')\to0$. This holds in particular if $\inf_c\sup_t|N_{\mu_n}(t)-N_{\mu_n'}(t+c)|=o(n)$, or if $\Wone(\mu_n,\mu_n')=o(n)$. The same statement holds with \emph{benign} replaced by \emph{tempered with limit $v_\infty$} or by \emph{catastrophic}. If, in addition, the bias $\bfn(\Sigma_n,\theta_n^\star)\to0$ and $\Gamma_{c_n}\to0$ (\cref{thm:bias}) along the shifts $c_n$ with $\Psi_{c_n}=o(n)$, then also $\bfn(\Sigma'_n,\theta'^\star_n)\to0$, so the excess risk of the perturbed problem vanishes.
\end{corollary}

The most useful special case concerns low-rank changes, such as the spikes created by feature learning \citep{ba2022high,moniri2024theory,dandi2024two} or by removing or adding a few features.

\begin{corollary}[Rank-$k$ updates of arbitrary size]\label{cor:rankk}
Let $\Sigma$ and $\Sigma'=\Sigma+E$ be positive semi-definite with ranks larger than $n$, where $E$ is symmetric with $\mathrm{rank}(E)\le k$. Then
\[
|\vf(\Sigma')-\vf(\Sigma)|\le\frac{3k}{n},\qquad\text{and, if } k\le\tfrac{n(1-\vf)}{6},\qquad\Big|\log\frac{\kappa(\Sigma')}{\kappa(\Sigma)}\Big|\le\frac{6k}{n(1-\vf)},
\]
independently of $\|E\|$.
\end{corollary}
By Weyl's inequalities \citep{horn2012matrix}, the counting functions of $\Sigma$ and $\Sigma'$ differ by at most $k$ at every level, so the claims follow from \cref{thm:local} with $c=0$. \Cref{thm:lower}(b) shows that $3k/n$ cannot be improved beyond a factor $3$.

\paragraph{Uniform multiplicative perturbations and the bias.} When all eigenvalues move by a bounded factor, a cruder argument controls ratios, including the bias and the full risk, and allows $p=\infty$ in the trace-class case.

\begin{proposition}[Uniform log-spectral perturbations]\label{prop:winf}
Let $(t_i,m_i)_{i\le p}$ and $(t'_i,m'_i)_{i\le p}$ with $p>n$ (or $p=\infty$ with $\sum_ie^{t_i}<\infty$ and $\sum_im_i<\infty$) satisfy $t'_i=t_i+c+\Delta_i$ and $m'_i=m_ie^{\eta_i}$, where $|\Delta_i|\le\delta$ and $|\eta_i|\le\eta$. Then $|u'-u-c|\le\delta$, $e^{-4\delta}\le\vf'/\vf\le e^{4\delta}$, and $e^{-4\delta-\eta}\le\bfn'/\bfn\le e^{4\delta+\eta}$. Consequently, if $e^{4\delta}\vf<1$,
\[
e^{-4\delta-\eta}(1-\vf)\;\le\;\frac{\Rde_n'}{\Rde_n}\;\le\;e^{4\delta+\eta}\,\frac{1-\vf}{1-e^{4\delta}\vf}.
\]
\end{proposition}

\paragraph{The bias in the same geometry.} The bias $\bfn=\int(1-\sig(t-u))^2\,\dd\nu(t)$ admits the same treatment, with the counting function replaced by the \emph{cumulative signal function} $M_\nu(t)=\nu((-\infty,t])$, the signal energy carried by eigenvalues at most $e^{t}$. Integration by parts gives $\bfn=\int M_\nu(t)\,\chi(t-u)\,\dd t$ with $\chi=2(1-\sig)\sig'$, a kernel with $\int\chi=1$, $\max\chi=\frac{8}{27}$ and exponential decay away from the threshold. For signal measures $\nu,\nu'$ and a shift $c$, let
\[
\Gamma_c:=\int_\R\big|M_\nu(t)-M_{\nu'}(t+c)\big|\,\chi(t-u)\,\dd t .
\]

\begin{theorem}[Localized transport stability of the bias]\label{thm:bias}
Let $\bfn,\bfn'$ be the bias functionals of $(\mu,\nu)$ and $(\mu',\nu')$, and let $\varepsilon_u=|\log\kappa(\mu')-c-\log\kappa(\mu)|$. Then
\[
|\bfn'-\bfn|\;\le\;\Gamma_c+\big(e^{2\varepsilon_u}-1\big)\big(\bfn+\Gamma_c\big).
\]
In particular, if $\Lambda_c\le n(1-\vf)/6$, then $\varepsilon_u\le6\Lambda_c/(n(1-\vf))$ by \cref{thm:local}(ii), and the right-hand side depends only on localized discrepancies.
\end{theorem}

Combining \cref{thm:local,thm:bias} with the identity $\Rde_n'-\Rde_n=\big[(\sigma^2+\bfn)(\vf'-\vf)+(1-\vf)(\bfn'-\bfn)\big]/\big((1-\vf)(1-\vf')\big)$ bounds the change of the full risk. The bias lives on signal \emph{below} the threshold ($1-\sig\approx1$), but $\chi$ charges only signal mass that is transported \emph{across} the threshold window. Mass moved far below the threshold, where $(1-\sig)^2\approx1$ throughout, costs almost nothing. A coupling-based variant with exponential moments is given in \cref{prop:bias} (\cref{app:proofs}).

\paragraph{Extensions: ridge regression and infinite dimensions.}
\begin{proposition}\label{prop:ext}
\textbf{(a) Ridge.} For a penalty $\lambda>0$ (estimator $(X^\top X+n\lambda I)^{-1}X^\top y$), the deterministic equivalent is \eqref{eq:de} with $\kappa=\kappa_\lambda$ the unique solution of $\sum_i\lambda_i/(\lambda_i+\kappa)=n(1-\lambda/\kappa)$ \citep{cheng2024dimension,bach2024high,misiakiewicz2024nonasymptotic}, which exists for every $p\ge1$. \Cref{thm:local}(i),(ii), \cref{cor:rankk} and \cref{thm:bias} hold verbatim with $c=0$ (and with $\lambda$ replaced by $e^{-c}\lambda$ in the shifted problem for $c\ne0$).
\textbf{(b) Infinite dimensions.} Let $\Sigma,\Sigma'$ be trace-class with infinitely many positive eigenvalues (or $\lambda>0$), and $\|\theta^\star\|_\Sigma,\|\theta'^\star\|_{\Sigma'}<\infty$. Then \cref{thm:local,thm:bias} hold whenever their right-hand sides are finite (with strict inequality in the hypothesis of \cref{thm:local}(ii)), and \cref{cor:rankk} holds for compact self-adjoint perturbations of rank at most $k$.
\end{proposition}

\begin{remark}[Transfer to the random risk]\label{rem:transfer}
Let $\hat{\mathcal R},\hat{\mathcal R}'$ be the excess risks of the estimators for the two problems. Suppose an approximation theorem gives $|\hat{\mathcal R}-\Rde_n|\le\epsilon$ and $|\hat{\mathcal R}'-\Rde_n'|\le\epsilon'$, each with probability at least $1-\delta$. Then, with probability at least $1-2\delta$, $|\hat{\mathcal R}'-\hat{\mathcal R}|\le|\Rde_n'-\Rde_n|+\epsilon+\epsilon'$, and $|\Rde_n'-\Rde_n|$ is controlled by \cref{thm:local,thm:bias}. The non-asymptotic theorems cited in \cref{sec:setting} supply $\epsilon,\epsilon'$ under their respective assumptions.
\end{remark}

\section{The transport scale is necessary}
\label{sec:lower}

\begin{theorem}[Lower bounds]\label{thm:lower}
\textbf{(a) Norm and spectral-distribution closeness do not preserve benignness.} For $n\ge2$, let $P_n$ be the orthogonal projection onto $n$ of the last $n^2$ coordinates in $\R^{1+n^2}$, and set
\[
\Sigma_n=\mathrm{diag}\big(1,\,n^{-2}I_{n^2}\big),\qquad \Sigma_n'=\Sigma_n+(n^{-1}-n^{-2})P_n .
\]
Then $\|\Sigma_n-\Sigma_n'\|_{\op}<1/n$ and $\|\Sigma_n-\Sigma_n'\|_F<n^{-1/2}$, while $\|\Sigma_n\|_{\op}=1$ and $\|\Sigma_n\|_F\ge1$. The empirical spectral distributions of $\Sigma_n,\Sigma'_n$ are within Kolmogorov distance $1/n$, and their log-spectra are within (dimension-normalized) $1$-Wasserstein distance $\log n/n$. Nevertheless,
\[
\vf(\Sigma_n)\le\frac{4}{n}\to0,\qquad \vf(\Sigma_n')\to\varphi^{-4}=\tfrac{7-3\sqrt5}{2},\qquad V(\Sigma_n')\to\frac{\sigma^2}{\varphi^4-1}=\frac{2\sigma^2}{5+3\sqrt5}\approx0.171\,\sigma^2,
\]
where $\varphi=(1+\sqrt5)/2$. Thus a sequence of variance-benign problems is mapped to a tempered one.

\textbf{(b) The rate in \cref{cor:rankk} is attained up to a factor $3$.} For $\alpha\in(0,1)$ and $k=\lfloor\alpha n\rfloor$, let $\Sigma_n=n^{-2}I_{n^2}$ and $\Sigma_n''=\Sigma_n+(1-n^{-2})Q_k$, where $Q_k$ is a coordinate projection of rank $k$. Then $\vf(\Sigma''_n)-\vf(\Sigma_n)\to\alpha$, while $k/n\to\alpha$.

\textbf{(c) The constant in \cref{cor:w1} is attained,} as stated there.
\end{theorem}

Part (a) isolates the conceptual point. Classical random-matrix theory compares spectra through their \emph{empirical spectral distributions}, which normalize by the dimension $p$ \citep{marchenko1967distribution,bai2010spectral}. In the proportional regime $p\asymp n$ this is equivalent to our normalization by $n$. Benign overfitting, however, lives in the regime $p\gg n$, and there only the $n$-normalized transport cost is meaningful. Here $n$ eigenvalues, a vanishing fraction $1/n$ of the spectrum, are moved by $\log n$ on the log scale. This costs $n\log n$ in $\Wone$ and $n$ in counting distance, which is exactly at the scale where \cref{thm:local} permits an $O(1)$ change. The golden ratio arises because the limiting threshold equation is $\frac{1}{1+K}+\frac1K=1$.

\section{Experiments}
\label{sec:exp}

All experiments use the exact deterministic equivalent (DE) and Monte Carlo simulation of $\hat\theta=X^{+}y$. Details, additional tables and the validation of the DE are in \cref{app:exp}. The code reproduces every number in this paper in a few minutes on a laptop CPU.

\begin{figure}[t]
\centering
\includegraphics[width=\linewidth]{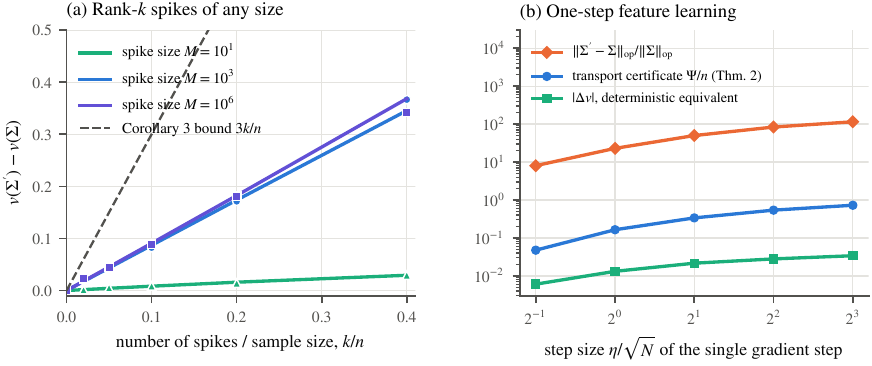}
\vspace{-1.5em}
\caption{\textbf{(a)} Starting from a benign isotropic covariance ($n=200$, $p=4000$, $\vf=0.05$), we replace $k$ eigenvalues by $M\in\{10,10^3,10^6\}$. Lines: DE; markers: simulation ($12$ draws). The change in $\vf$ saturates in $M$ and stays below the bound $3k/n$ of \cref{cor:rankk}, although the operator norm of the perturbation grows to $10^6$. \textbf{(b)} One giant gradient step on the first layer of a two-layer $\tanh$ network ($d=128$, $N=1024$ neurons, $n=256$). The feature covariance changes by up to $115\times$ its operator norm, whereas the localized transport certificate $\inf_c\Psi_c/n$ of \cref{thm:local} remains below $1$ and tracks the actual change in $\vf$.}
\label{fig:main}
\vspace{-0.8em}
\end{figure}

\paragraph{Operator norm versus transport (\cref{fig:intro}b, \cref{fig:main}a).} The perturbation of \cref{thm:lower}(a) has operator norm below $1/n$, yet both the DE and the simulated interpolator converge to $\vf\approx\varphi^{-4}$ (at $n=240$, $p=57{,}600$: simulated $0.151$, DE $0.148$, limit $0.146$). In the opposite direction, spikes of size up to $10^6$ change $\vf$ by roughly $k/n$, comfortably within $3k/n$. Simulation and DE agree to within two Monte Carlo standard errors at every point.

\paragraph{One-step feature learning (\cref{fig:main}b).} Following \citet{ba2022high}, we take a two-layer network $f(x)=a^\top\tanh(Wx)$ with $W_0$ Gaussian, $x\sim\mathcal N(0,I_d)$, and a single-index target. We take one gradient step $W_1=W_0-\eta\nabla_WL$ with $\eta\propto\sqrt N$. The gradient is nearly rank one (ratio of its top two singular values $0.019$), and the resulting feature covariance develops a large outlier, up to $115\times$ the original operator norm. We then fit pure noise with the minimum-norm interpolator on fresh samples in feature space, which isolates the variance term. Across step sizes, the DE value of $\vf$ moves by at most $0.034$, and the transport certificate $\inf_c\Psi_c/n$ upper-bounds it within a factor between $8$ and $22$. Because the counting functions differ also in the bulk, $\sup_t|N-N'|$ is large ($140$--$300$). The localized and translation-reduced discrepancy is what keeps the certificate informative. For step sizes up to $4\sqrt N$, simulations agree with the DE ($|\Delta|\le0.034$; \cref{tab:fl}). At the largest step, $8\sqrt N$, the saturated $\tanh$ features are far from Gaussian. The simulated $\vf$ then deviates from the DE ($0.652$ vs.\ $0.496$), a failure of the DE approximation rather than of the stability theory.

\begin{table}[t]
\centering
\caption{Random ReLU features ($3000$) of the $1797$ \texttt{digits} images, with the population taken as the uniform distribution over all images ($n=300$, $\log\kappa=-1.56$). Spectral surgeries act linearly on the real features in the population eigenbasis. ``rel.\ op'' is $\|\Sigma'-\Sigma\|_{\op}/\|\Sigma\|_{\op}$; $\Psi_0/n$ is the certificate of \cref{thm:local} (with $c=0$); $\vf$ (sim.) is computed from $60$ independent training sets.}
\label{tab:digits}
\small
\begin{tabular}{@{}lrrrcc@{}}
\toprule
Perturbation & rank & rel.\ op & $\Psi_0/n$ & $\vf$ (DE) & $\vf$ (sim.) \\
\midrule
none (original features) & 0 & $0$ & 0.000 & 0.527 & 0.560 $\pm$ 0.006 \\
flatten top-5 eigenvalues & 5 & $0.609$ & 0.000 & 0.527 & 0.554 $\pm$ 0.005 \\
flatten top-20 eigenvalues & 20 & $0.883$ & 0.002 & 0.526 & 0.534 $\pm$ 0.003 \\
flatten top-60 eigenvalues & 60 & $0.983$ & 0.055 & 0.509 & 0.483 $\pm$ 0.002 \\
amplify top-5 by $10^3$ & 5 & $999$ & 0.000 & 0.527 & 0.562 $\pm$ 0.007 \\
amplify top-20 by $10^3$ & 20 & $999$ & 0.002 & 0.527 & 0.775 $\pm$ 0.015 \\
amplify 300 sub-threshold dirs.\ by 10 & 300 & $2.5\times 10^{-3}$ & 0.962 & 0.465 & 0.470 $\pm$ 0.004 \\
amplify 300 sub-threshold dirs.\ by 100 & 300 & $0.0278$ & 2.411 & 0.558 & 0.556 $\pm$ 0.003 \\
amplify 300 sub-threshold dirs.\ by 1000 & 300 & $0.281$ & 2.785 & 0.730 & 0.727 $\pm$ 0.003 \\
\bottomrule
\end{tabular}
\vspace{-0.6em}
\end{table}

\paragraph{Real features (\cref{tab:digits}).} We take random ReLU features of the \texttt{digits} images and apply spectral surgeries. Amplifying the top five directions by $10^3$ (relative operator change $999$) leaves $\vf$ unchanged in both DE and simulation. Amplifying $300=n$ sub-threshold directions by a factor $10^3$, a relative operator change of only $0.28$, raises $\vf$ from $0.53$ to $0.73$, with DE and simulation in agreement. This is the real-data analogue of \cref{thm:lower}(a), and the certificate $\Psi_0/n$ correctly flags it. Two rows deserve comment. On the unperturbed features the DE underestimates the simulated $\vf$ by about $6\%$, as expected for structured non-Gaussian data \citep{wei2022more,loureiro2021learning}. When the top $20$ directions are amplified by $10^3$, the simulated $\vf$ jumps to $0.78$, while the DE predicts no change. In that case the log-spectrum near the threshold is essentially unchanged ($\Psi_0/n=0.002$), so \emph{no} functional of the spectrum can predict the jump. It reflects dependence between high- and low-variance coordinates of real data, which lies outside the Gaussian-equivalent theory (see \cref{sec:discussion}).

\section{Related work}
\label{sec:related}

\textbf{Benign overfitting.} Beyond the linear theory of \citet{bartlett2020benign,tsigler2023benign}, related analyses cover uniform convergence and norm bounds \citep{koehler2021uniform}, classification \citep{chatterji2021finite}, kernels \citep{liang2020just,rakhlin2019consistency,haas2023mind,mallinar2022benign}, and neural networks \citep{frei2022benign,cao2022benign}. \citet{shamir2022implicit} shows that benign overfitting depends delicately on the problem. Our results quantify this sensitivity and identify the geometry in which it disappears.
\textbf{Deterministic equivalents and spectral theories.} The functional \eqref{eq:de} and its relatives are derived in \citet{dobriban2018high,hastie2022surprises,bach2024high,cheng2024dimension,misiakiewicz2024nonasymptotic,defilippis2024dimension} and appear as learning curves in \citet{bordelon2020spectrum,canatar2021spectral,jacot2020kernel,simon2023eigenlearning,atanasov2024scaling}. See also \citet{advani2020high,mei2022generalization,hu2023universality}. These works compute the risk for a given spectrum. We study how it varies across spectra, in a geometry adapted to the threshold.
\textbf{Spikes and feature learning.} Outliers in covariance spectra are classical \citep{baik2005phase}. One gradient step creates them in neural representations \citep{ba2022high,moniri2024theory,dandi2024two}. \Cref{cor:rankk} shows that such low-rank changes cannot by themselves turn benign interpolation into harmful interpolation unless their rank is proportional to $n$.
\textbf{Optimal transport.} We use only one-dimensional optimal transport \citep{villani2009optimal,santambrogio2015optimal,peyre2019computational}. Its role here is structural: monotone couplings of log-eigenvalues, costs localized by the logistic kernel, and translation reduction encoding scale invariance.

\section{Discussion and limitations}
\label{sec:discussion}

We have shown that benign overfitting is a continuous property in a threshold-localized transport geometry on log-spectra with cost scale $n$, and that it is discontinuous in the operator-norm and spectral-distribution topologies. Two consequences seem practically relevant. First, robustness certificates for interpolating models can be phrased as transport budgets: rank-$o(n)$ changes, or changes confined far from the threshold, are harmless. Second, the threshold-localized sensitivity distribution $\pi$ identifies which eigen-directions of a representation actually control noise fitting.

\textbf{Limitations.} (1) Our theorems concern the deterministic equivalent. Transfer to the random risk (\cref{rem:transfer}) inherits the assumptions of the approximation theorem used, and these can fail for structured data, as \cref{tab:digits} shows for one surgery. (2) The constants in \cref{thm:local,thm:bias} are explicit but not optimized. Only the Wasserstein constant $8/27$ and the rate $k/n$ are shown to be sharp. (3) The first-variation formula (\cref{thm:firstvar}) is stated for ridgeless regression. With $\lambda>0$, the denominator of $\pi_j$ gains the term $n\lambda/\kappa$, and we have not pursued this. (4) Extending the geometry to classification and to the full training dynamics of feature learning is an exciting direction.

\section*{Reproducibility statement}
All theorems are proved in \cref{app:proofs}. All experiments are specified in \cref{sec:exp} and \cref{app:exp}, including dimensions, seeds and numbers of repetitions. They rely only on \texttt{numpy}, \texttt{scipy} and the \texttt{digits} dataset shipped with \texttt{scikit-learn}. Every inequality in \cref{sec:firstvar,sec:global} was additionally stress-tested on random instances, and every first-variation formula was checked against finite differences (\cref{app:checks}).

\bibliography{references}
\bibliographystyle{iclr2026_conference}

\newpage
\appendix

\section{Proofs}
\label{app:proofs}

\subsection{Elementary facts about the logistic function}

\begin{lemma}\label{lem:logistic}
For all $x,h\in\R$:
(i) $\sig(x+h)\le e^{|h|}\sig(x)$ and $1-\sig(x+h)\le e^{|h|}(1-\sig(x))$;
(ii) $\sig'(x+h)\ge e^{-|h|}\sig'(x)$;
(iii) $\sig'(x)=\frac{1}{4\cosh^2(x/2)}\le\min\{\frac14,e^{-|x|}\}$;
(iv) $\max_{s\in[0,1],\,a\in[0,1]}s(1-s)|s-a|=\frac4{27}$, attained exactly at $(s,a)=(\frac23,0)$ and $(\frac13,1)$;
(v) $\psi=2\sig'(1+\sig)$ satisfies $\int\psi=3$ and $\max\psi=\frac{4}{3\sqrt3}$.
\end{lemma}
\begin{proof}
(i) For $h\ge0$, $\sig(x+h)/\sig(x)=(1+e^{-x})/(1+e^{-x}e^{-h})\le e^{h}$. For $h<0$, $\sig$ is increasing, so the ratio is at most $1$. The second claim follows from $1-\sig(x)=\sig(-x)$.
(ii) Since $\cosh(y+z)\le\cosh(y)e^{|z|}$, we get $\cosh^2((x+h)/2)\le\cosh^2(x/2)e^{|h|}$. Now use (iii).
(iii) The identity is direct, and $4\cosh^2(x/2)\ge\max\{4,e^{|x|}\}$.
(iv) $s(1-s)|s-a|\le s(1-s)\max\{s,1-s\}$, and $\max_s s^2(1-s)=4/27$ is attained at $s=2/3$. By symmetry the same holds with $s$ and $1-s$ exchanged. Equality requires $a\in\{0,1\}$.
(v) $\int2\sig'=2$ and $\int2\sig\sig'=[\sig^2]_{-\infty}^{\infty}=1$. Writing $\psi=2\sig(1-\sig^2)$ as a function of $\sig\in(0,1)$, it is maximized at $\sig=1/\sqrt3$ with value $\frac{4}{3\sqrt3}$.
\end{proof}

\subsection{Proof of \cref{prop:logspec}}
(i) $F_\mu(w):=\sum_i\sig(t_i-w)$ is continuous and strictly decreasing, with limits $p$ at $-\infty$ and $0$ at $+\infty$. Since $p>n$, the equation $F_\mu(u)=n$ has a unique solution. This solution is $\log\kappa$ by \eqref{eq:kappa}. Dividing by $n$ shows that $P_\mu$ has total mass $1$.
(ii) We have $\lambda_i^2/(\lambda_i+\kappa)^2=\sig_i^2$ and $\kappa^2\lambda_i\theta_i^2/(\lambda_i+\kappa)^2=m_i(1-\sig_i)^2$, so $B=\bfn/(1-\vf)$ and $V=\sigma^2\vf/(1-\vf)$.
(iii) Shifting all $t_i$ by $c$ shifts the solution of (i) by $c$ and leaves all $\sig_i$ unchanged. Under $\theta^\star\mapsto e^{-c/2}\theta^\star$, the masses $m_i=\lambda_i\theta_i^2$ are unchanged.
(iv) Since $N_\mu(t)=\sum_i\mathbf 1\{t<t_i\}$, Fubini's theorem gives $\int N_\mu(t)g'(t-w)\,\dd t=\sum_i\int_{-\infty}^{t_i}g'(t-w)\,\dd t=\sum_i g(t_i-w)$ for $g\in\{\sig,\sig^2\}$, because $g(-\infty)=0$. \qed

\subsection{Proof of \cref{thm:firstvar}}
Let $\Phi(t,u)=\sum_i\sig(t_i-u)-n$. Then $\partial_u\Phi=-\sum_i\sig_i'<0$. Since $\Phi$ is real-analytic, the implicit function theorem shows that $u(t)$ is real-analytic, and hence so are $\vf,\bfn,\Rde_n$ (note that $\vf<1$ because $\sum\sig_i^2<\sum\sig_i=n$). Write $S_1=\sum_i\sig_i'=n(1-\vf)$.

(a) Differentiating $\Phi(t,u(t))=0$ in $t_j$ gives $\sig_j'-S_1\,\partial_{t_j}u=0$.

(b) $n\,\partial_{t_j}\vf=\partial_{t_j}\sum_i\sig_i^2=2\sig_j\sig_j'-\sum_i2\sig_i\sig_i'\,\partial_{t_j}u=2\sig_j'\big(\sig_j-\sum_i\pi_i\sig_i\big)$.

(c) $\partial_{t_j}\bfn=-2m_j(1-\sig_j)\sig_j'+\sum_i2m_i(1-\sig_i)\sig_i'\,\partial_{t_j}u=2\sig_j'\big(\sum_i\pi_im_i(1-\sig_i)-m_j(1-\sig_j)\big)$.

(d) This is immediate since $\bfn$ is linear in $m$ and $u$ does not depend on $m$.

The formula for $\partial_{t_j}\Rde_n$ follows from the quotient rule applied to $\Rde_n=(\sigma^2\vf+\bfn)/(1-\vf)$. Summing (a) over $j$ gives $\sum_j\pi_j=1$. Summing (b) gives $\frac2n(\sum_j\sig_j'\sig_j-S_1\Ep[\sig])=0$, and (c) is analogous. For \eqref{eq:gradbound}, $|\sig_j-\Ep[\sig]|\le1$ together with \cref{lem:logistic}(iii) gives the first inequality. \Cref{lem:logistic}(iv), applied with $a=\Ep[\sig]\in[0,1]$, gives the second. \qed

\subsection{Proof of \cref{cor:w1}}
\emph{Upper bound.} Fix $c\in\R$ and let $\Delta_i=t'_{(i)}-c-t_{(i)}$. Consider the path $t(\tau)=(t_{(i)}+\tau\Delta_i)_{i\le p}$ for $\tau\in[0,1]$. Each $t(\tau)$ has $p>n$ atoms, so \cref{thm:firstvar} applies along the whole path, and $\vf(t(1))=\vf(\mu')$ by translation invariance. Hence
\[
|\vf(\mu')-\vf(\mu)|=\Big|\int_0^1\sum_j\partial_{t_j}\vf(t(\tau))\,\Delta_j\,\dd\tau\Big|\le\frac{8}{27n}\sum_j|\Delta_j| .
\]
Taking the infimum over $c$ proves the bound.

\emph{Sharpness.} Fix $n\ge2$, $N=n^2$ and $\sig_0=(n-\frac23)/N$. Let $\mu$ consist of one atom at $t_1=\log2$ and $N$ atoms at $t_0=\log(\sig_0/(1-\sig_0))$. Then $\sig(t_1)=\frac23$ and $\sig(t_0)=\sig_0$, so $\sig(t_1)+N\sig(t_0)=n$ and $u=0$. With $\sig_1'=\frac29$,
\[
\Ep[\sig]=\frac{\frac23\cdot\frac29+N\sig_0^2(1-\sig_0)}{\frac29+N\sig_0(1-\sig_0)}\le\frac{\frac4{27}+1}{(n-\frac23)(1-\frac1n)}\xrightarrow[n\to\infty]{}0 ,
\]
using $N\sig_0^2\le1$ and $\sig_0\le1/n$. By \cref{thm:firstvar}(b), $n\,\partial_{t_1}\vf=\frac49(\frac23-\Ep[\sig])\to\frac{8}{27}$. Let $\mu_\varepsilon$ be obtained by moving $t_1$ to $t_1+\varepsilon$. Then $\Wone(\mu,\mu_\varepsilon)=\inf_c(|\varepsilon-c|+N|c|)=\varepsilon$, so $n|\vf(\mu_\varepsilon)-\vf(\mu)|/\Wone(\mu,\mu_\varepsilon)\to n|\partial_{t_1}\vf|$ as $\varepsilon\to0$. Letting $n\to\infty$ proves that the supremum is $8/27$. \qed

\subsection{Proof of \cref{thm:local}}
Let $\tilde\mu$ be $\mu'$ shifted by $-c$, i.e.\ with atoms $t'_i-c$. Then $N_{\tilde\mu}(t)=N_{\mu'}(t+c)$, $\vf(\tilde\mu)=\vf'$, and $\tilde u:=\log\kappa(\tilde\mu)=\log\kappa(\mu')-c$ by \cref{prop:logspec}(iii). Write $D:=|N_\mu-N_{\tilde\mu}|$, $F_\rho(w)=\sum_{t\in\rho}\sig(t-w)$ and $G_\rho(w)=\sum_{t\in\rho}\sig(t-w)^2$ for $\rho\in\{\mu,\tilde\mu\}$, where sums run over atoms with multiplicity.

\emph{(i).} We have $n(\vf'-\vf)=G_{\tilde\mu}(\tilde u)-G_\mu(u)=[G_{\tilde\mu}(\tilde u)-G_{\tilde\mu}(u)]+[G_{\tilde\mu}(u)-G_\mu(u)]$. By \cref{prop:logspec}(iv), the second bracket is at most $\int D\,(\sig^2)'(\cdot-u)=\int D\cdot2\sig\sig'(\cdot-u)$ in absolute value. For the first bracket, $G_{\tilde\mu}'(w)=-2\sum\sig\sig'$ and $F_{\tilde\mu}'(w)=-\sum\sig'<0$, so $|G_{\tilde\mu}'|\le2|F_{\tilde\mu}'|$ pointwise. Integrating between $u$ and $\tilde u$ and using the monotonicity of $F_{\tilde\mu}$,
\[
|G_{\tilde\mu}(\tilde u)-G_{\tilde\mu}(u)|\le2|F_{\tilde\mu}(\tilde u)-F_{\tilde\mu}(u)|=2|F_\mu(u)-F_{\tilde\mu}(u)|\le2\int D\,\sig'(\cdot-u),
\]
where we used $F_{\tilde\mu}(\tilde u)=n=F_\mu(u)$ and \cref{prop:logspec}(iv). Adding the two bounds gives $n|\vf'-\vf|\le\int D\,(2\sig'+2\sig\sig')(\cdot-u)=\Psi_c$.

\emph{(ii).} Let $g(w)=F_{\tilde\mu}(w)-n$. It is strictly decreasing with $g(\tilde u)=0$, and $|g(u)|=|F_{\tilde\mu}(u)-F_\mu(u)|\le\Lambda_c$. Its slope satisfies $-g'(w)=S(w):=\sum_{t\in\tilde\mu}\sig'(t-w)$. At $w=u$,
\[
S(u)=F_{\tilde\mu}(u)-G_{\tilde\mu}(u)\ge(n-\Lambda_c)-\Big(n\vf+\int D\,2\sig\sig'(\cdot-u)\Big)\ge n(1-\vf)-3\Lambda_c\ge\tfrac12n(1-\vf),
\]
using $2\sig\sig'\le2\sig'$ and the hypothesis. By \cref{lem:logistic}(ii), $S(w)\ge e^{-|w-u|}S(u)$. Set $x=\Lambda_c/S(u)\le2\Lambda_c/(n(1-\vf))\le\frac13$. If $x=0$, then $D=0$ almost everywhere, so $\tilde\mu=\mu$ and $\tilde u=u$. Otherwise let $a=3x\le1$. Then
\[
g(u+a)\le\Lambda_c-\int_u^{u+a}S\le S(u)\big(x-3xe^{-3x}\big)<0,
\]
because $e^{-3x}\ge e^{-1}>\frac13$. Symmetrically $g(u-a)>0$. Hence $|\tilde u-u|<a\le6\Lambda_c/(n(1-\vf))$.

\emph{(iii).} The bounds follow from Hölder's inequality and \cref{lem:logistic}(iii),(v). \qed

\subsection{Proofs of \cref{cor:benign,cor:rankk}}
\Cref{cor:benign} is immediate from \cref{thm:local}(i),(iii), since $|\vf(\Sigma_n')-\vf(\Sigma_n)|\le\inf_c\Psi_c/n\to0$.

For \cref{cor:rankk}, let $\lambda_1(A)\ge\lambda_2(A)\ge\cdots$ denote the eigenvalues of a symmetric matrix $A$. Since $E$ has at most $k$ positive eigenvalues, $\lambda_{k+1}(E)\le0$. Weyl's inequality $\lambda_{i+k}(\Sigma+E)\le\lambda_i(\Sigma)+\lambda_{k+1}(E)$ \citep[Thm.~4.3.1]{horn2012matrix} therefore gives $\lambda_{i+k}(\Sigma')\le\lambda_i(\Sigma)$. Symmetrically, applied to $\Sigma=\Sigma'+(-E)$, it gives $\lambda_{i+k}(\Sigma)\le\lambda_i(\Sigma')$. Hence, for every $x>0$, the number of eigenvalues larger than $x$ differs between $\Sigma$ and $\Sigma'$ by at most $k$, i.e.\ $\sup_t|N_\mu(t)-N_{\mu'}(t)|\le k$. Only positive eigenvalues enter $\mu,\mu'$, and both have more than $n$ atoms by assumption. \Cref{thm:local} with $c=0$ gives $\Psi_0\le3k$ and $\Lambda_0\le k$. \qed

\subsection{Proof of \cref{thm:bias}}
Let $\tilde\nu$ be $\nu'$ shifted by $-c$, so $M_{\tilde\nu}(t)=M_{\nu'}(t+c)$, and let $\tilde u=\log\kappa(\mu')-c$, so that $\bfn'=\bfn(\tilde\nu,\tilde u)$ by \cref{prop:logspec}(iii). Here $\bfn(\rho,w):=\int(1-\sig(t-w))^2\,\dd\rho(t)$. Since $(1-\sig)^2$ decreases from $1$ to $0$, Fubini's theorem gives $\bfn(\rho,w)=\int M_\rho(t)\,\chi(t-w)\,\dd t$ with $\chi=-\frac{\dd}{\dd x}(1-\sig)^2=2(1-\sig)\sig'$, exactly as in \cref{prop:logspec}(iv). Hence $|\bfn(\tilde\nu,u)-\bfn(\nu,u)|\le\Gamma_c$. By \cref{lem:logistic}(i), every term of $\bfn(\tilde\nu,\cdot)$ changes by a factor in $[e^{-2|\tilde u-u|},e^{2|\tilde u-u|}]$ when the threshold moves from $u$ to $\tilde u$. Therefore $|\bfn(\tilde\nu,\tilde u)-\bfn(\tilde\nu,u)|\le(e^{2\varepsilon_u}-1)\bfn(\tilde\nu,u)\le(e^{2\varepsilon_u}-1)(\bfn+\Gamma_c)$. Adding the two bounds proves the claim. The values $\int\chi=1$ and $\max\chi=\max_{\sig}2\sig(1-\sig)^2=\frac{8}{27}$ (at $\sig=\frac13$) are elementary. The identity for $\Rde_n'-\Rde_n$ follows by expanding $(\sigma^2\vf'+\bfn')(1-\vf)-(\sigma^2\vf+\bfn)(1-\vf')$. \qed

\subsection{Proof of \cref{prop:ext}}
\emph{(a) Ridge.} Let $H_\rho(w)=F_\rho(w)-n+n\lambda e^{-w}$. It is continuous and strictly decreasing, tends to $+\infty$ as $w\to-\infty$ and to $-n$ as $w\to+\infty$, so it has a unique root $u=\log\kappa_\lambda$ for every $p\ge1$. Moreover $\sum_i\sig_i^2<\sum_i\sig_i=n(1-\lambda/\kappa_\lambda)\le n$, so $\vf<1$. We take $c=0$ and write $\tilde\mu=\mu'$ and $\tilde u=u'$. In the proof of \cref{thm:local}(i), the only place where the threshold equation enters is the bound on $|F_{\tilde\mu}(\tilde u)-F_{\tilde\mu}(u)|$. Now $H_{\tilde\mu}(\tilde u)=0$ and $H_{\tilde\mu}(u)=F_{\tilde\mu}(u)-F_\mu(u)$ (because $H_\mu(u)=0$). The two summands of $H_{\tilde\mu}(\tilde u)-H_{\tilde\mu}(u)=[F_{\tilde\mu}(\tilde u)-F_{\tilde\mu}(u)]+n\lambda(e^{-\tilde u}-e^{-u})$ have the same sign, since both functions are decreasing. Hence $|F_{\tilde\mu}(\tilde u)-F_{\tilde\mu}(u)|\le|F_{\tilde\mu}(u)-F_\mu(u)|\le\Lambda_0$, and the rest of the proof of (i) is unchanged. For (ii), apply the argument to $g=H_{\tilde\mu}$, whose slope is $S(w)+n\lambda e^{-w}$. At $w=u$,
\[
S(u)+\tfrac{n\lambda}{\kappa_\lambda}\ge\big(n(1-\tfrac{\lambda}{\kappa_\lambda})-\Lambda_0\big)-\big(n\vf+2\Lambda_0\big)+\tfrac{n\lambda}{\kappa_\lambda}=n(1-\vf)-3\Lambda_0 .
\]
Since $n\lambda e^{-w}\ge e^{-|w-u|}n\lambda e^{-u}$, the slope again satisfies $-g'(w)\ge e^{-|w-u|}(-g'(u))$, and the proof of (ii) goes through verbatim. \Cref{cor:rankk} follows as before, and \cref{thm:bias} uses only (ii) and \cref{lem:logistic}. For $c\neq0$, scaling $(\Sigma',\lambda)\mapsto(e^{-c}\Sigma',e^{-c}\lambda)$ leaves the ridge problem invariant, which reduces to the case $c=0$.

\emph{(b) Infinite dimensions.} Order the eigenvalues decreasingly and let $\mu_P$, $\nu_P$ keep the $P$ largest. Then $N_{\mu_P}=\min(N_\mu,P)$, so $|N_{\mu_P}-N_{\mu'_P}|\le|N_\mu-N_{\mu'}|$ (and similarly after a shift by $c$), and $M_{\nu_P}\to M_\nu$ pointwise with $|M_{\nu_P}|\le\nu(\R)<\infty$. Since $\sig(t_i-w)\le\lambda_ie^{-w}$ and $\sum_i\lambda_i<\infty$, dominated convergence shows $F_{\mu_P}\to F_\mu$, $G_{\mu_P}\to G_\mu$ and $\bfn(\nu_P,\cdot)\to\bfn(\nu,\cdot)$ locally uniformly. $F_\mu$ is continuous and strictly decreasing, with $F_\mu(-\infty)=\infty$ (infinitely many atoms) or with the ridge term. Hence $F_{\mu_P}(u\pm\epsilon)\to F_\mu(u\pm\epsilon)$, which lie on either side of $n$, and so $u_P\to u$. Consequently $\vf(\mu_P)\to\vf(\mu)$ and $\bfn(\nu_P,u_P)\to\bfn(\nu,u)$. For the discrepancies, $\psi(x)\le4\sig'(x)\le4e^{-|x|}$ and $e^{-|t-u_P|}\le e^{\epsilon}e^{-|t-u|}$ for $|u_P-u|\le\epsilon$, while $\psi(x)\ge2\sig'(x)\ge\frac12e^{-|x|}$. Therefore $\Psi_c<\infty$ provides an integrable dominating function, and $\Psi_c(\mu_P,\mu'_P)\to\Psi_c(\mu,\mu')$, and likewise for $\Lambda_c$. For $\Gamma_c$, the integrand is bounded and $\chi$ is integrable. Applying the finite-dimensional results for each $P$ and letting $P\to\infty$ proves (b). For compact self-adjoint operators, the min--max principle gives the eigenvalue inequalities used in the proof of \cref{cor:rankk}. \qed

\subsection{A coupling form of the bias bound}
\begin{proposition}[Bias under a coupling]\label{prop:bias}
Let $(t_i,m_i)_{i\le p}$ and $(t'_i,m'_i)_{i\le p}$, write $t'_i=t_i+c+\Delta_i$ and $m'_i=m_ie^{\eta_i}$, and let $\varepsilon_u=|u'-u-c|$. Then
\[
e^{-2\varepsilon_u}\,\E_{w}\big[e^{-2|\Delta|-|\eta|}\big]\;\le\;\frac{\bfn'}{\bfn}\;\le\;e^{2\varepsilon_u}\,\E_{w}\big[e^{2|\Delta|+|\eta|}\big],
\]
where $\E_w$ averages over directions with the bias-share distribution $w$ of \eqref{eq:dists}.
\end{proposition}

\subsection{Proofs of \cref{prop:winf,prop:bias}}
Let $\tilde t_i=t'_i-c=t_i+\Delta_i$, so that $\tilde u=u'-c$ is the threshold of $(\tilde t_i)$.

\emph{Threshold (\cref{prop:winf}).} Since $\tilde t_i\le t_i+\delta$ and $\sig$ is increasing, $F_{\tilde\mu}(w)\le F_\mu(w-\delta)$, so $F_{\tilde\mu}(u+\delta)\le F_\mu(u)=n=F_{\tilde\mu}(\tilde u)$. As $F_{\tilde\mu}$ is strictly decreasing, $\tilde u\le u+\delta$. The bound $\tilde u\ge u-\delta$ follows in the same way.

\emph{Ratios.} Let $x_i=t_i-u$ and $\tilde x_i=\tilde t_i-\tilde u$. Then $|\tilde x_i-x_i|\le|\Delta_i|+|\tilde u-u|$. By \cref{lem:logistic}(i), $\sig(\tilde x_i)^2/\sig(x_i)^2$ and $(1-\sig(\tilde x_i))^2/(1-\sig(x_i))^2$ both lie in $[e^{-2|\tilde x_i-x_i|},e^{2|\tilde x_i-x_i|}]$. Under the hypotheses of \cref{prop:winf}, $|\tilde x_i-x_i|\le2\delta$. Averaging the first ratio over $i$ with weights $\sig_i^2/(n\vf)$ gives the bound on $\vf'/\vf$. Averaging $e^{\eta_i}$ times the second ratio over $w$ gives the bound on $\bfn'/\bfn$, because $\bfn'/\bfn=\sum_iw_ie^{\eta_i}(1-\sig(\tilde x_i))^2/(1-\sig(x_i))^2$. This identity, with $|\tilde x_i-x_i|\le|\Delta_i|+\varepsilon_u$, also proves \cref{prop:bias}.

\emph{Risk.} We have $\Rde_n'=(\sigma^2\vf'+\bfn')/(1-\vf')\le e^{4\delta+\eta}(\sigma^2\vf+\bfn)/(1-e^{4\delta}\vf)$, which gives the upper bound. Similarly, $\Rde_n'\ge e^{-4\delta-\eta}(\sigma^2\vf+\bfn)/(1-e^{-4\delta}\vf)\ge e^{-4\delta-\eta}(\sigma^2\vf+\bfn)$, which gives the lower bound after dividing by $\Rde_n=(\sigma^2\vf+\bfn)/(1-\vf)$. \qed

\subsection{Proof of \cref{thm:lower}}
\emph{(a).} \emph{Isotropic part.} For $\Sigma^{\rm iso}=n^{-2}I_{n^2}$, \eqref{eq:kappa} reads $n^2\cdot n^{-2}/(n^{-2}+\kappa)=n$, so $\kappa=n^{-1}-n^{-2}$, $\sig_i=1/n$ and $\vf=n^2\cdot n^{-2}/n=1/n$.

\emph{Perturbed part.} $\Sigma^{\rm pert}$ has $n$ eigenvalues $n^{-1}$ and $n^2-n$ eigenvalues $n^{-2}$. Writing $\kappa=K/n$, the learnabilities are $1/(1+K)$ and $1/(1+nK)$, and \eqref{eq:kappa} becomes $h_n(K):=\frac{1}{1+K}+\frac{n-1}{1+nK}=1$. Each $h_n$ is strictly decreasing on $(0,\infty)$, with $h_n(0^+)=n>1$ and $h_n(\infty)=0$, so the root $K_n$ is unique. For $K>0$, $\big|\frac{n-1}{1+nK}-\frac1K\big|=\frac{1+K}{K(1+nK)}\le\frac{1+K}{nK^2}$. Hence $h_n\to h$, with $h(K)=\frac1{1+K}+\frac1K$, uniformly on compact subsets of $(0,\infty)$. The function $h$ is strictly decreasing, and $h(\varphi)=\varphi^{-2}+\varphi^{-1}=1$ because $\varphi^2=\varphi+1$. For any $\epsilon\in(0,\varphi)$, $h(\varphi-\epsilon)>1>h(\varphi+\epsilon)$, so uniform convergence on $[\varphi-\epsilon,\varphi+\epsilon]$ gives $K_n\in(\varphi-\epsilon,\varphi+\epsilon)$ for all large $n$. Thus $K_n\to\varphi$, and
\[
\vf(\Sigma^{\rm pert})=\frac{1}{(1+K_n)^2}+\frac{n-1}{(1+nK_n)^2}\to\frac1{(1+\varphi)^2}=\varphi^{-4},
\]
since the second term is at most $1/(nK_n^2)\to0$.

\emph{Adding the unit spike.} $\Sigma_n=\mathrm{diag}(1)\oplus\Sigma^{\rm iso}$ is a rank-one update of $0\oplus\Sigma^{\rm iso}$, whose counting measure equals that of $\Sigma^{\rm iso}$. By \cref{cor:rankk}, $|\vf(\Sigma_n)-1/n|\le3/n$, and likewise $|\vf(\Sigma_n')-\vf(\Sigma^{\rm pert})|\le3/n$. The limits of $V=\sigma^2\vf/(1-\vf)$ follow, using $\varphi^4=(7+3\sqrt5)/2$.

\emph{Norms and distances.} $\Sigma_n'-\Sigma_n$ has $n$ nonzero eigenvalues equal to $n^{-1}-n^{-2}$. This gives the operator and Frobenius norms. The two spectra differ in $n$ of $1+n^2$ eigenvalues, so the Kolmogorov distance between the empirical spectral distributions is $n/(1+n^2)<1/n$. The monotone coupling of the log-spectra moves $n$ atoms by $\log n$, so the dimension-normalized $1$-Wasserstein distance is $n\log n/(1+n^2)<\log n/n$.

\emph{(b).} Write $\kappa=K/n$. There are $k$ eigenvalues equal to $1$, with learnability $1/(1+K/n)$, and $n^2-k$ equal to $n^{-2}$, with learnability $1/(1+nK)$. Dividing \eqref{eq:kappa} by $n$ gives $h_n(K)=\frac{k/n}{1+K/n}+\frac{n-k/n}{1+nK}=1$, and $h_n\to h(K)=\alpha+1/K$ uniformly on compact subsets of $(0,\infty)$. The argument of (a) gives $K_n\to1/(1-\alpha)$, and then $\vf=\frac1n\big[\frac{k}{(1+K_n/n)^2}+\frac{n^2-k}{(1+nK_n)^2}\big]\to\alpha$. Since $\vf(\Sigma_n)=1/n$, the claim follows. Moreover $\Sigma_n''-\Sigma_n$ has rank $k$.

\emph{(c)} was proved with \cref{cor:w1}. \qed

\subsection{Relation to effective ranks}
\begin{lemma}\label{lem:effrank}
Let $k=\#\{i:\lambda_i\ge\kappa\}$, $T=\sum_{i>k}\lambda_i$ and $S=\sum_{i>k}\lambda_i^2$. If $k\le n/2$, then
\[
\tfrac14\Big(\frac kn+\frac{nS}{4T^2}\Big)\;\le\;\vf\;\le\;\frac kn+\frac{4nS}{T^2}.
\]
\end{lemma}
\begin{proof}
For $i\le k$, $\sig_i\in[\frac12,1)$. For $i>k$, $\lambda_i/(2\kappa)\le\sig_i\le\lambda_i/\kappa$. From $n=\sum_{i\le k}\sig_i+\sum_{i>k}\sig_i$ we get $\sum_{i>k}\sig_i\in[n-k,n-k/2]$, and also $\sum_{i>k}\sig_i\in[T/(2\kappa),T/\kappa]$. Hence $\kappa\in[T/(2n),2T/n]$ when $k\le n/2$. Therefore $n\vf=\sum_{i\le k}\sig_i^2+\sum_{i>k}\sig_i^2$ lies between $\frac k4+\frac{S}{4\kappa^2}\ge\frac k4+\frac{n^2S}{16T^2}$ and $k+\frac{S}{\kappa^2}\le k+\frac{4n^2S}{T^2}$. Dividing by $n$ gives the claim.
\end{proof}

\section{Experimental details and additional results}
\label{app:exp}

\paragraph{Computing the DE.} We solve $\sum_i\sig(t_i-u)=n$ by bisection to a tolerance of $10^{-13}$, followed by Newton polishing. All functionals are then explicit.

\paragraph{Simulation.} For a diagonal $\Sigma$ (without loss of generality), we draw $X=Z\Sigma^{1/2}$ with i.i.d.\ standard Gaussian $Z$. We compute $\hat\theta=X^\top(XX^\top)^{-1}y$, and the excess risk $\|\hat\theta-\theta^\star\|_\Sigma^2$ in closed form. For variance-only experiments we set $\theta^\star=0$ and $\sigma=1$, so the risk equals $V$, and we report $\bar V/(1+\bar V)$, which equals $\vf$ under the DE. Standard errors of $\vf$ are propagated from those of $\bar V$ by the delta method.

\paragraph{Validation of the DE.} \Cref{tab:val} compares the DE \eqref{eq:de} with simulated excess risk for $n=150$, $p\approx3000$, $\sigma=0.5$, and $\theta^\star$ with i.i.d.\ $\mathcal N(0,9/p)$ entries ($100$ draws).

\begin{table}[h]
\centering
\caption{Deterministic equivalent versus simulation of the minimum-norm interpolator.}
\label{tab:val}
\small
\begin{tabular}{@{}lccccc@{}}
\toprule
Spectrum & $\vf$ & $B$ (DE) & $V$ (DE) & risk (DE) & risk (sim.) \\
\midrule
$\lambda_i=i^{-1/2}$ & 0.080 & 0.2894 & 0.0217 & 0.3111 & 0.3109 $\pm$ 0.0006 \\
$\lambda_i=i^{-1}$ & 0.217 & 0.0148 & 0.0692 & 0.0840 & 0.0863 $\pm$ 0.0013 \\
$\lambda_i=i^{-3/2}$ & 0.380 & 0.0008 & 0.1531 & 0.1538 & 0.1541 $\pm$ 0.0025 \\
5 spikes $+$ flat tail & 0.079 & 0.0089 & 0.0214 & 0.0303 & 0.0296 $\pm$ 0.0006 \\
\bottomrule
\end{tabular}
\end{table}

\Cref{tab:val2} extends the check to Rademacher designs ($x=\Sigma^{1/2}z$ with i.i.d.\ uniform signs $z$) and to ridge penalties $\lambda\in\{10^{-3},10^{-2}\}$ (estimator $(X^\top X+n\lambda I)^{-1}X^\top y$), with the same $n,\sigma,\theta^\star$ and $100$ draws. The largest relative deviation between DE and simulation is $4.2\%$.

\begin{table}[h]
\centering
\caption{Deterministic equivalent (with $\kappa=\kappa_\lambda$, \cref{prop:ext}(a)) versus simulation for Gaussian and Rademacher designs, ridgeless and ridge.}
\label{tab:val2}
\small
\begin{tabular}{@{}llccc@{}}
\toprule
Spectrum & design & $\lambda$ & risk (DE) & risk (sim.) \\
\midrule
$\lambda_i=i^{-1}$ & Gaussian & 0 & 0.0822 & 0.0798 $\pm$ 0.0011 \\
$\lambda_i=i^{-1}$ & Gaussian & 0.001 & 0.0787 & 0.0804 $\pm$ 0.0011 \\
$\lambda_i=i^{-1}$ & Gaussian & 0.01 & 0.0584 & 0.0584 $\pm$ 0.0008 \\
$\lambda_i=i^{-1}$ & Rademacher & 0 & 0.0822 & 0.0811 $\pm$ 0.0010 \\
$\lambda_i=i^{-1}$ & Rademacher & 0.001 & 0.0787 & 0.0787 $\pm$ 0.0011 \\
$\lambda_i=i^{-1}$ & Rademacher & 0.01 & 0.0584 & 0.0585 $\pm$ 0.0008 \\
5 spikes $+$ flat tail & Gaussian & 0 & 0.0302 & 0.0300 $\pm$ 0.0005 \\
5 spikes $+$ flat tail & Gaussian & 0.001 & 0.0288 & 0.0277 $\pm$ 0.0005 \\
5 spikes $+$ flat tail & Gaussian & 0.01 & 0.0225 & 0.0222 $\pm$ 0.0005 \\
5 spikes $+$ flat tail & Rademacher & 0 & 0.0302 & 0.0305 $\pm$ 0.0006 \\
5 spikes $+$ flat tail & Rademacher & 0.001 & 0.0288 & 0.0294 $\pm$ 0.0006 \\
5 spikes $+$ flat tail & Rademacher & 0.01 & 0.0225 & 0.0217 $\pm$ 0.0006 \\
\bottomrule
\end{tabular}
\end{table}

\paragraph{\Cref{fig:intro}b.} We use $n\in\{10,20,40,80,160,240\}$ with $p=n^2$ and $30$ draws per point, and $\sigma=1$, $\theta^\star=0$. DE curves are computed up to $n=10^4$.

\paragraph{\Cref{fig:main}a.} We use $n=200$, $p=4000$ and $\Sigma=I$ ($\vf=0.05$). We replace $k\in\{0,4,10,20,40,80\}$ eigenvalues by $M\in\{10,10^3,10^6\}$, with $12$ draws per point.

\paragraph{\Cref{fig:main}b and \cref{tab:fl}.} Inputs are $x\sim\mathcal N(0,I_{128})$. The network has $N=1024$ neurons, $W_0$ with i.i.d.\ $\mathcal N(0,1/d)$ entries, $a$ with i.i.d.\ $\mathcal N(0,1/N)$ entries, and $\tanh$ activation. The target is $y=\mathrm{ReLU}(z)-(2\pi)^{-1/2}+z/2+0.1\,\xi$ with $z=\langle\beta,x\rangle$ and $\|\beta\|=1$. One full-batch gradient step on $W$ for the squared loss over $4096$ samples uses step size $\eta=c\sqrt N$, $c\in\{0.5,1,2,4,8\}$. Feature covariances are estimated from $24{,}000$ fresh inputs. The variance is measured by fitting pure noise on $n=256$ fresh samples ($40$ draws). $\Psi_c$ is minimized over a grid of $81$ shifts $c\in[-1,1]$. $\sup_t|N-N'|$ is reported for $c=0$.

\begin{table}[h]
\centering
\caption{One-step feature learning. Columns: step size $c=\eta/\sqrt N$; top eigenvalue of the feature covariance; relative operator change; $\sup_t|N-N'|$ at $c=0$; certificate $\inf_c\Psi_c/n$; $\vf$ by DE and by simulation.}
\label{tab:fl}
\small
\begin{tabular}{@{}rrrrccc@{}}
\toprule
$c$ & $\lambda_{\max}$ & rel.\ op & $\sup|N-N'|$ & $\Psi/n$ & $\vf$ (DE) & $\vf$ (sim.) \\
\midrule
0 & 5.3 & 0.00 & 0 & 0.000 & 0.530 & 0.524 $\pm$ 0.005 \\
0.5 & 43.2 & 8.03 & 140 & 0.048 & 0.524 & 0.523 $\pm$ 0.005 \\
1 & 122.7 & 22.93 & 255 & 0.165 & 0.517 & 0.523 $\pm$ 0.005 \\
2 & 266.8 & 49.91 & 299 & 0.338 & 0.509 & 0.517 $\pm$ 0.005 \\
4 & 446.7 & 83.60 & 269 & 0.540 & 0.502 & 0.536 $\pm$ 0.005 \\
8 & 614.9 & 115.11 & 230 & 0.727 & 0.496 & 0.652 $\pm$ 0.007 \\
\bottomrule
\end{tabular}
\end{table}

\paragraph{\Cref{tab:digits}.} The inputs are the $1797$ \texttt{digits} images ($8\times8$), standardized per pixel and scaled by $1/8$. Features are $\mathrm{ReLU}(Wx)$ with $W\in\R^{3000\times64}$ i.i.d.\ standard Gaussian, centred over the dataset. The population is the uniform distribution over the $1797$ images, so the population covariance is exact and has rank $1796$. We express the features in its eigenbasis. ``Flatten top-$k$'' rescales the top $k$ coordinates so that their variances equal $\lambda_{k+1}$. ``Amplify'' multiplies the variances of the indicated coordinates by the stated factor. The $300$ sub-threshold directions are the first $300$ eigen-directions with $\log\lambda_i<\log\kappa-2$. Training sets of size $n=300$ are drawn without replacement ($60$ draws). We verified that the anomalous row is not a numerical artefact: solving via $(XX^\top)^{-1}$ and via least squares agree to ten digits, and the Gram matrices have condition numbers of about $5\cdot10^6$.

\subsection{Numerical verification of the theory}
\label{app:checks}
(1) The derivative formulas of \cref{thm:firstvar} agree with central finite differences, up to floating-point error, at $150$ random (instance, coordinate) pairs from $30$ random instances with $n\in[5,60]$, $p\le n+400$ and mixed spectra. (2) Over $400$ random pairs $(\mu,\mu')$, covering small dense perturbations, arbitrary rank-$k$ moves, different cardinalities and shifted spectra, the largest observed ratios of the left- to the right-hand sides were $0.54$ for \cref{thm:local}(i), $0.18$ for \cref{thm:local}(ii), $0.39$ for \cref{cor:w1}, $0.49$ for \cref{prop:winf} and $0.21$ for \cref{cor:rankk}. The bias bounds of \cref{prop:bias} held on all $300$ random instances tested. Over $600$ further random pairs, half of them with a ridge penalty $\lambda>0$ and including changes of signal and of dimension, the largest observed ratios were $0.50$ for \cref{thm:local}(i), $0.19$ for \cref{thm:local}(ii) and $0.96$ for \cref{thm:bias}. (3) For the construction in the proof of \cref{cor:w1}, $n\,\partial_{t_1}\vf$ equals $0.2912,0.2958,0.2961$ for $n=100,1000,3000$ (limit $8/27\approx0.2963$). (4) For \cref{thm:lower}(a), $\vf(\Sigma^{\rm pert})=0.1519,0.1465,0.1460,0.1459$ for $n=10^2,\dots,10^5$ (limit $0.14590$).

\section{Use of large language models}
A large language model was used to assist with drafting and editing the text, with deriving and checking proofs, and with writing experiment code. All authors take full responsibility for the content of this paper.

\end{document}

%% file: math_commands.tex
\usepackage{amsmath,amsfonts,bm}

\def\eqref#1{equation~\ref{#1}}

\def\1{\bm{1}}

\def\vf{{\bm{f}}}

\DeclareMathAlphabet{\mathsfit}{\encodingdefault}{\sfdefault}{m}{sl}
\SetMathAlphabet{\mathsfit}{bold}{\encodingdefault}{\sfdefault}{bx}{n}

\newcommand{\E}{\mathbb{E}}

\newcommand{\R}{\mathbb{R}}

